\documentclass[sigconf, screen]{acmart}

\usepackage{graphicx}
\graphicspath{{figures/}{samples/figures/}}
\usepackage{multicol}
\usepackage{multirow}
\usepackage{booktabs}
\usepackage{makecell}
\usepackage{gradient-text}
\usepackage{xspace}
\usepackage{bbding}
\usepackage[table]{xcolor}
\newcommand{\best}[1]{\cellcolor{blue!25}\textbf{#1}}
\newcommand{\second}[1]{\cellcolor{blue!10}#1}

\renewcommand\footnotetextcopyrightpermission[1]{}
\AtBeginDocument{%
  }

\setcopyright{acmlicensed}
\copyrightyear{2018}
\acmYear{2018}
\acmDOI{XXXXXXX.XXXXXXX}
\acmConference[Conference acronym 'XX]{Make sure to enter the correct
  conference title from your rights confirmation email}{June 03--05,
  2018}{Woodstock, NY}
\acmISBN{978-1-4503-XXXX-X/2018/06}

\acmSubmissionID{7707}

\begin{document}

\title{Weather-Conditioned Branch Routing for Robust LiDAR-Radar 3D Object Detection}


\author{Hongsheng Li}
\authornote{Both authors contributed equally to this research.}

\affiliation{%
  \institution{Shenzhen International Graduate School, Tsinghua University}
  \city{Shenzhen}
  \state{Guangdong}
  \country{China}
  }
\email{lihs25@mails.tsinghua.edu.cn}

\author{Lingfeng Zhang}

\authornotemark[1]
\affiliation{%
  \institution{Shenzhen International Graduate School, Tsinghua University}
  \city{Shenzhen}
  \state{Guangdong}
  \country{China}
  }
\email{lfzhang715@gmail.com}

\author{Zexian Yang}

\affiliation{%
  \institution{College of Computer and Data Science, Fuzhou University}
  \city{Fuzhou}
  \state{Fujian}
  \country{China}
  }
\email{yangzexian@fzu.edu.cn}

\author{Liang Li}

\affiliation{%
  \institution{Institute of Information Engineering, CAS}
  \city{Beijing}
  \country{China}
  }
\email{liliang@iie.ac.cn}

\author{Rong Yin}

\affiliation{%
  \institution{Institute of Information Engineering, CAS}
    \city{Beijing}
  \country{China}
  }
\email{yinrong@iie.ac.cn}

\author{Xiaoshuai Hao}
\authornote{Corresponding Authors.}
\authornote{Project Leader.}
\affiliation{%
  \institution{Xiaomi EV}
  \city{Beijing}
  \country{China}}
\email{haoxiaoshuai@xiaomi.com}

\author{Wenbo Ding}
\authornotemark[2]
\affiliation{%
    \institution{Shenzhen International Graduate School, Tsinghua University}
  \city{Shenzhen}
  \state{Guangdong}
  \country{China}}
\email{ding.wenbo@tsinghua.edu.cn}

\renewcommand{\shortauthors}{Trovato et al.}

\begin{abstract}

 Robust 3D object detection in adverse weather is highly challenging due to the varying reliability of different sensors. While existing LiDAR-4D radar fusion methods improve robustness, they predominantly rely on fixed or weakly adaptive pipelines, failing to dynamically adjust modality preferences as environmental conditions change. To bridge this gap,  we reformulate multi-modal perception as a weather-conditioned branch routing problem. Instead of computing a single fused output, our framework explicitly maintains three parallel 3D feature streams: a pure LiDAR branch, a pure 4D radar branch, and a condition-gated fusion branch. Guided by a condition token extracted from visual and semantic prompts, a lightweight router dynamically predicts sample-specific weights to softly aggregate these representations. Furthermore, to prevent branch collapse, we introduce a weather-supervised learning strategy with auxiliary classification and diversity regularization to enforce distinct, condition-dependent routing behaviors. Extensive experiments on the K-Radar benchmark demonstrate that our method achieves state-of-the-art performance. Furthermore, it provides explicit and highly interpretable insights into modality preferences, transparently revealing how adaptive routing robustly shifts reliance between LiDAR and 4D radar across diverse adverse-weather scenarios. The source code with be released.
\end{abstract}

\begin{teaserfigure}
  \includegraphics[width=\textwidth]{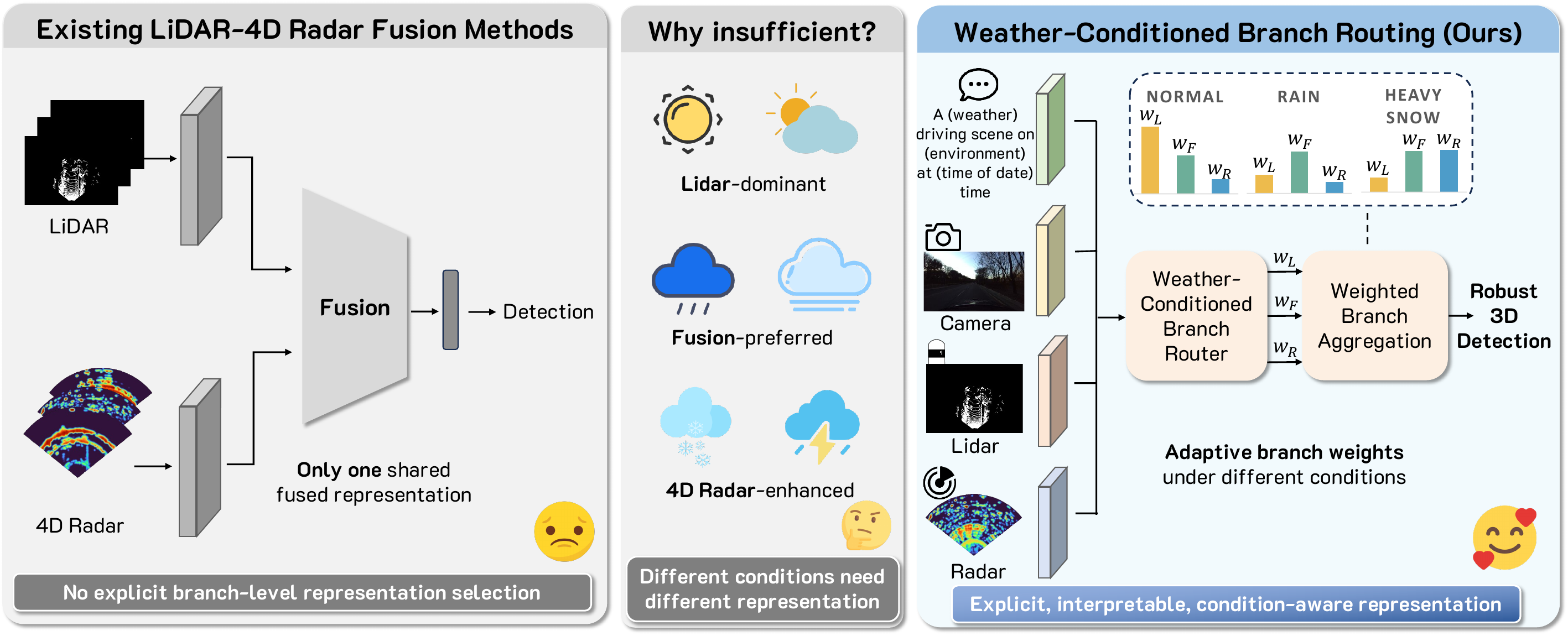}
  \caption{Comparison between existing LiDAR-4D radar fusion methods and our proposed approach. Existing methods rely on a single, fixed fused representation, which struggles to adapt to changing environments. In contrast, our weather-conditioned branch routing explicitly maintains multiple representation branches and dynamically assigns adaptive weights based on the current weather condition.}
  \label{fig:teaser}
\end{teaserfigure}


\maketitle

\begin{figure*}[t]
    \centering
    \includegraphics[width=\textwidth]{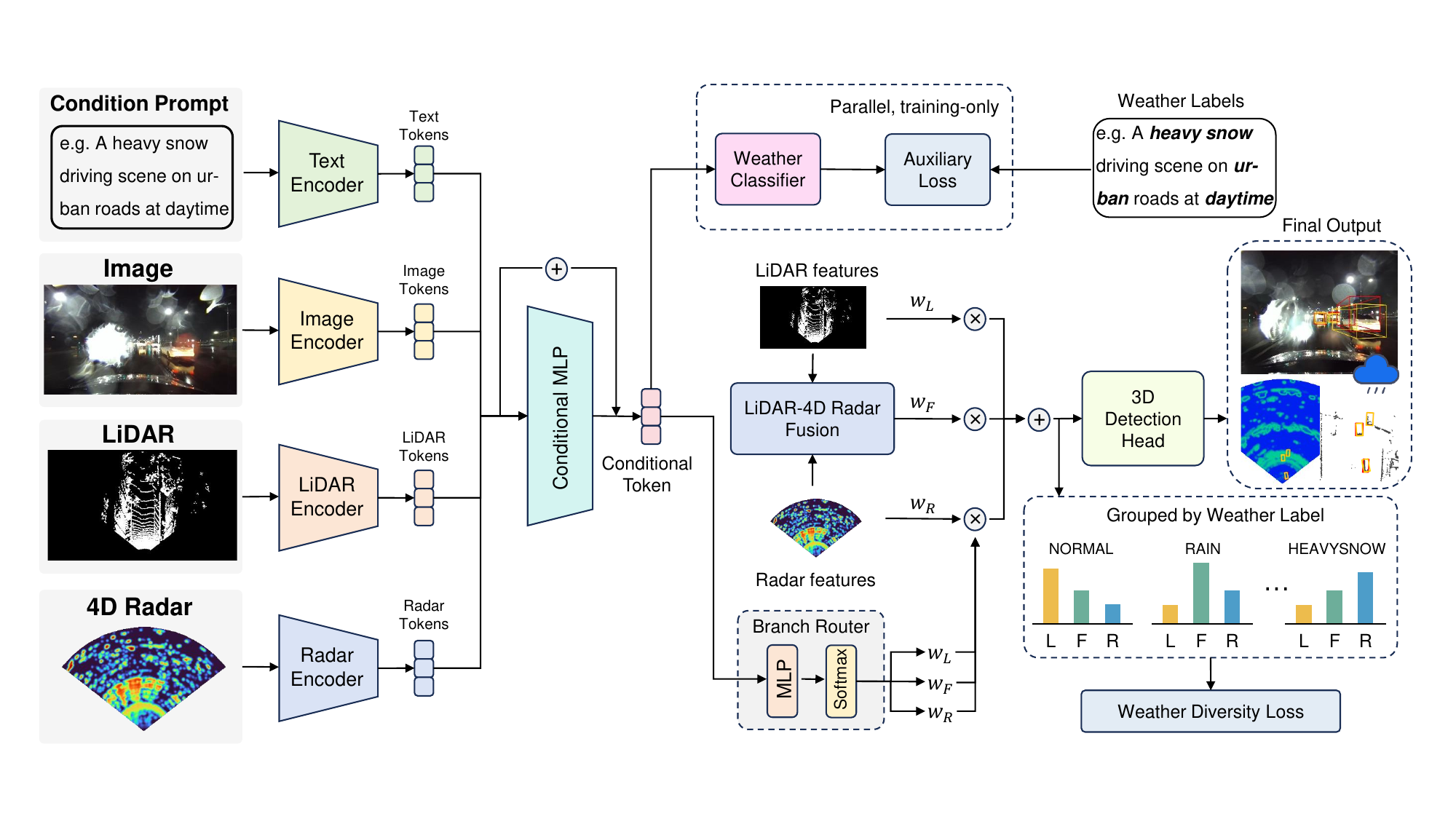}
    \vspace{-3em}
    \caption{Framework overview of the proposed weather-conditioned branch routing method. The model takes LiDAR, 4D radar, and camera images as inputs. First, a condition token is generated by fusing visual features and semantic text prompts, which is further refined by sensor context. Next, LiDAR and radar features are processed through three parallel streams: a LiDAR-only branch, a radar-only branch, and a condition-gated fusion branch. A branch router then utilizes the condition token to predict adaptive weights ($w_L, w_F, w_R$). These weights softly aggregate the branch-wise BEV features for the final 3D detection head. During training, auxiliary weather classification and diversity losses are applied to ensure meaningful routing behavior.}
    \Description{}
    \label{fig:framework}
    \vspace{-1.5em}
\end{figure*}

\section{Introduction}
Multi-sensor perception is essential for reliable autonomous driving, especially in adverse weather where the strengths and weaknesses of different sensors become more pronounced.\cite{Bijelic2020seeing, song2024robustness,hao2025mimo,hao2025msc,hao2025safemap,zhang2025mapnav,zhang2025nava,tang2025roboafford,hao2025roboafford++,zhang2025your,zhang2025socialnav,zhang2026you,zhang2025video,xiao2025team,zhang2025team,kong2026robosense,hao2025really,shan2025stability}. LiDAR provides accurate geometric structure but may degrade under severe weather, while radar is generally more robust yet less spatially precise. This complementarity has motivated a growing body of LiDAR-radar fusion methods, many of which have achieved strong and even state-of-the-art performance on challenging benchmarks such as K-Radar\cite{paek2022kradar,wang2022interfusion,chae20243dlrf,huang2025l4dr,zheng2025railway,zhang2024trihelper,zhang2025multi,gong2025stairway,zhang2025humanoidpano,wu2025evaluating,liu2025toponav,li2025vquala,cheng2025exploring,zhang2025lips,hao2026h2r,fu2026sef,zhang2026meshmimic}.

Despite their effectiveness, most existing methods are still built upon fixed or unified fusion pipelines, where sensor interaction is largely determined by the architecture itself\cite{liu2023bevfusion,bai2022transfusion,zhao2025unibevfusion}. Such designs are powerful, but they do not explicitly address an important property of adverse-weather perception: the most reliable representation may vary with the weather condition. In some scenarios, LiDAR-dominant geometry may remain the most informative signal, while in others radar cues or a stronger fused representation may be more favorable. This observation suggests that adverse-weather fusion should be formulated not only as feature fusion, but also as conditional representation selection.
Moreover, when the fusion behavior is fully implicit, it becomes difficult to interpret why a model succeeds or fails under particular weather regimes.For safety-critical deployment, this lack of interpretability can be as limiting as raw accuracy.

In this work, we revisit LiDAR-radar fusion from this perspective and formulate it as a weather-conditioned branch routing problem. Instead of relying on a single fixed fusion path, we explicitly maintain three candidate representations: a LiDAR-only branch, a Radar-only branch, and a fused LiDAR-radar branch. A condition-aware router then predicts adaptive weights over these branches according to the current scene context. Importantly, weather information is not used as a hard rule for branch assignment. Instead, it is encoded into a learned condition token, which guides routing in a data-driven manner. This design yields a routing-based fusion framework that is conceptually different from conventional fixed-fusion methods\cite{liu2023bevfusion,bai2022transfusion,li2022deepfusion} and complementary to more heavily redesigned unified-fusion architectures\cite{zhao2025unibevfusion,yin2024fusion,chen2023futr3d}.
By decoupling candidate representations from the routing decision, the framework provides both flexibility and transparency: it can adapt branch preference online while preserving analyzable branch semantics.

A practical challenge of this formulation is optimization. In our preliminary observations, naïvely training the router often leads to branch collapse, where the model quickly over-relies on a dominant fused branch and fails to learn meaningful weather-dependent preferences. To address this issue, we introduce a weather-supervised routing learning scheme. Specifically, we couple the routing process with an auxiliary weather classification objective, which encourages the condition token to preserve discriminative weather information, and a weather diversity regularization term, which promotes distinguishable branch preference patterns across weather conditions. Together, these objectives improve the learnability and discriminability of weather-conditioned routing.

We evaluate the proposed framework on the K-Radar dataset\cite{paek2022kradar}. Experimental results show that our method achieves competitive performance against strong LiDAR-4D radar baselines and demonstrates state-of-the-art-level effectiveness within the branch-routing formulation. More importantly, beyond accuracy alone, our framework provides interpretable weather-aware branch preference patterns, enabling a clearer analysis of how modality reliance changes across adverse-weather conditions. We believe this explicit routing perspective offers a valuable and practical alternative to fixed fusion pipelines for robust multi-sensor 3D perception.

Our main contributions are as follows:
\begin{itemize}
    \item We propose a weather-conditioned branch routing framework for LiDAR-radar 3D object detection, which reformulates adverse-weather fusion as adaptive weighting over LiDAR-only, Radar-only, and fused feature branches.
    \item  We introduce a weather-supervised routing learning strategy with auxiliary weather classification and weather diversity regularization to improve routing discriminability and robustness under adverse-weather conditions.

    \item We conduct extensive experiments and analysis on K-Radar, showing that the proposed method achieves competitive performance with strong baselines while providing interpretable weather-aware branch preference behaviors.
\end{itemize}

\section{Related works}
\textbf{Multi-modal 3D object detection.}
Multi-modal 3D object detection has been widely studied to improve perception robustness by combining complementary sensing modalities such as cameras, LiDAR, and radar\cite{sindagi2019mvx, vora2020pointpainting, li2022deepfusion, li2023logonet, huang2020epnet,hao2025mimo,hao2023mixgen,hao2024your,hao2025mapfusion,hao2024mapdistill,hao2024mbfusion}. Earlier studies mainly focused on LiDAR–camera fusion, and explored point-level, voxel-/pseudo-image-level, and BEV-level interaction strategies to enhance geometric and semantic representation\cite{sindagi2019mvx, vora2020pointpainting, li2023logonet,xia2024hinted,wu2023virtual,yan2023cross}. More recently, radar has been introduced as an additional sensing modality because of its robustness to adverse weather and long-range measurement capability\cite{paek2022kradar, Wu2023MVFusion,wang2023bi,wu2023virtual}. Existing radar-related methods typically perform fusion through early fusion, feature concatenation, cross-attention, or gated interaction modules\cite{wang2022interfusion, Song2024LiRaFusion, Wolters2025UnleashingHydra,li2025moe3d}, and have shown clear advantages over single-modality detectors in challenging environments. With the emergence of 4D radar, recent work further extends multi-modal detection from conventional radar–camera settings to LiDAR–4D radar fusion\cite{paek2022kradar, wang2022interfusion, chae20243dlrf,Qi2026FusionBevLA}, demonstrating that richer elevation and Doppler cues can provide additional benefits for 3D object detection. Nevertheless, most current multi-modal detectors still emphasize how to design a stronger unified fusion backbone, while less attention has been paid to explicitly modeling condition-dependent preference over different modality-specific or fused representations.

\textbf{Adverse-Weather 3D Object Detection.}
Robust 3D object detection under adverse weather has attracted increasing attention because sensor quality can change significantly across rain, snow, fog, and other challenging conditions\cite{Bijelic2020seeing,huang2024sunshine,song2024robustness,Qian2021RobustFoggy}. Existing efforts mainly improve robustness from the perspective of single-modality enhancement, such as adverse-weather simulation\cite{Hahner2021FogSO,Kong2023Robo3DTR}, LiDAR denoising\cite{chae20243dlrf}, domain adaptation\cite{Yang2021ST3D,wu2026unida3dunifieddomainadaptiveframework,xu2021spg,do2022lossdistillnet,wang2023ssda3d}, and knowledge distillation\cite{huang2024sunshine,Chae2024LiDARBasedA3}, with the goal of alleviating the degradation of LiDAR point clouds in corrupted environments. These methods improve detector generalization to some extent, but they still rely primarily on a single sensing stream and therefore remain limited when the dominant informative modality changes across weather conditions. Recent studies \cite{chae20243dlrf,huang2025l4dr}further emphasize that adverse-weather perception should account for the fact that different sensors exhibit different reliability under different weather, motivating the use of condition-aware mechanisms rather than treating all scenes with a fixed representation strategy. This observation suggests that robust 3D detection in adverse weather should not only enhance features under corruption, but also explicitly adapt representation preference according to the environmental condition.

\textbf{LiDAR–4D Radar Fusion for Robust 3D Detection.}
LiDAR–r-adar fusion has become an important direction for robust 3D perception because LiDAR provides accurate geometric structure while 4D radar offers stronger weather robustness, velocity cues, and longer sensing range\cite{paek2022kradar,kong2023rtnh+,xiong2023lxl}. Prior fusion methods, including InterFusion\cite{wang2022interfusion}, 3D-LRF\cite{chae20243dlrf}, and L4DR\cite{huang2025l4dr}, have shown that combining LiDAR and 4D radar can outperform single-modality detectors, especially under adverse weather. However, most existing approaches are still based on fixed or weakly adaptive fusion pipelines, where modality interaction is predefined by early fusion, feature-level fusion, attention, or gated enhancement modules. Although some methods introduce weather-aware gating or adaptive feature filtering, their adaptivity mainly operates within a unified fusion backbone and still aims to learn a single fused representation. In contrast, they do not explicitly model the possibility that different weather conditions may favor different representation branches, such as LiDAR-dominant, radar-dominant, or fused features. Therefore, existing LiDAR–4D radar fusion methods leave open an important question: how to explicitly perform weather-conditioned representation selection, rather than only weather-conditioned feature modulation, for robust multi-modal 3D object detection.

\vspace{-1em}

\section{Methods}

\subsection{Framework Overview}

The overall scheme of our framework is shown in Fig.~\ref{fig:framework}.
Our framework takes a LiDAR point cloud, a 4D radar sparse cube, and a front-view camera image as input.
Specifically, the LiDAR point cloud is $\mathbf{P}_{L} \in \mathbb{R}^{N_{L} \times 4}$, the 4D radar sparse cube is
$\mathbf{P}_{R} \in \mathbb{R}^{N_{R} \times 4}$, and the camera image is $\mathbf{I} \in \mathbb{R}^{H \times W \times 3}$,
where $N_{L}$ and $N_{R}$ are the numbers of LiDAR and radar voxel
points, and $H$, $W$ are the image height and width, respectively.
The camera image is first encoded by a lightweight 2D convolutional network pre-trained for weather classification, producing a visual condition token $\mathbf{c}_{v} \in \mathbb{R}^{512}$. 
This token is fused with a semantically grounded weather prompt token $\mathbf{c}_{p} \in \mathbb{R}^{512}$ from a frozen CLIP~\cite{radford2021clip} text encoder, and further refined by a sensor-aware mixing MLP that incorporates pooled radar and LiDAR context, yielding the final condition token $\hat{\mathbf{c}} \in \mathbb{R}^{512}$
.

$\mathbf{P}_{L}$ and $\mathbf{P}_{R}$ are each mapped through an
input layer to higher-dimensional voxel features
$\mathbf{L}^{0} \in \mathbb{R}^{N_{L} \times C_{0}}$ and
$\mathbf{R}^{0} \in \mathbb{R}^{N_{R} \times C_{0}}$,
and subsequently fed into three parallel three-layer sparse 3D
convolution networks.
At each layer $l \in \{1, 2, 3\}$, the \textit{LiDAR-only branch}
extracts voxel features
$\mathbf{L}^{l} \in \mathbb{R}^{N_{l} \times C_{l}}$,
the \textit{Radar-only branch} extracts
$\mathbf{R}^{l} \in \mathbb{R}^{M_{l} \times C_{l}}$,
and the \textit{fusion branch} produces
$\mathbf{F}^{l} \in \mathbb{R}^{N_{l} \times C_{l}}$
by augmenting the LiDAR features with condition-gated KNN attention over radar neighbors (see Sec.~\ref{sec:branch_routing}).
Each branch independently collapses its 3D voxel features to a BEV
feature map via sparse $z$-axis convolution and transposed convolution upsampling, and the per-layer BEV features are concatenated across all three layers to produce branch-wise BEV representations$\tilde{\mathbf{F}}_{L},\, \tilde{\mathbf{F}}_{R},\,
 \tilde{\mathbf{F}}_{F} \in \mathbb{R}^{B \times 768 \times 32 \times 180}$.
A weather-conditioned branch router predicts a soft routing weight
vector $\mathbf{w} = [w_{L},\, w_{R},\, w_{F}]$ from $\hat{\mathbf{c}}$ via a two-layer MLP, and assembles the final BEV feature $\tilde{\mathbf{F}} \in \mathbb{R}^{B \times 1536 \times 32 \times 180}$ as a weighted combination of the three branches (see Sec.~\ref{sec:branch_routing}).
$\tilde{\mathbf{F}}$ is then passed to an anchor-based detection head to produce the final 3D bounding box predictions.

\subsection{Weather-Conditioned Representation Modeling}

\label{sec:weather_condition}

The goal of this module is to construct a compact scene-level condition
token $\hat{\mathbf{c}} \in \mathbb{R}^{512}$ that captures the current
weather state and guides downstream branch routing.
The module combines two complementary sources of weather information.

\textbf{Visual condition token.}
The front-view camera image $\mathbf{I} \in \mathbb{R}^{3 \times H \times W}$
is passed through a lightweight three-layer 2D convolutional backbone
pre-trained for weather classification.
The backbone progressively downsamples the image with stride-4 convolutions,
producing a spatial feature map of shape $\mathbb{R}^{64 \times 11 \times 20}$,
which is then globally average-pooled and projected to obtain the visual
condition token $\mathbf{c}_{v} \in \mathbb{R}^{512}$.

\textbf{Semantic weather token.}
In parallel, we employ a frozen CLIP text encoder~\cite{radford2021clip}
to provide semantic weather priors.
A set of seven weather-class text prompts of the form
\textit{``A \{weather\} driving scene''}
(covering normal, overcast, fog, rain, sleet, light snow, and heavy snow)
are encoded offline into a weather vocabulary matrix
$\mathbf{W} \in \mathbb{R}^{7 \times 512}$, which is stored as a fixed buffer.
At runtime, the per-sample scene description prompt is encoded by the
same text encoder into a normalized prompt embedding
$\mathbf{p} \in \mathbb{R}^{512}$, and its soft alignment to each weather
class is computed as:
\begin{equation}
  \boldsymbol{\alpha} = \mathrm{softmax}(\mathbf{p}\,\mathbf{W}^{\top})
  \in \mathbb{R}^{7},
  \label{eq:weather_soft_attn}
\end{equation}
yielding a soft weather probability vector $\boldsymbol{\alpha}$.
The semantic weather token is then obtained as the attention-weighted
sum over the vocabulary,
$\mathbf{c}_{p} = \mathrm{Proj}(\boldsymbol{\alpha}\,\mathbf{W})
\in \mathbb{R}^{512}$,
where $\mathrm{Proj}$ is a linear projection followed by layer
normalization and ReLU.
This design encodes weather information as a \emph{continuous}
distribution over known weather classes rather than a hard label,
allowing the model to handle ambiguous or transitional weather conditions
in a data-driven manner.

\textbf{Sensor-aware condition token refinement.}
The visual and semantic tokens are averaged to form an initial condition
token $\mathbf{c} = \frac{1}{2}(\mathbf{c}_{v} + \mathbf{c}_{p})
\in \mathbb{R}^{512}$.
To further ground this token in the actual sensor observations of the
current scene, we extract lightweight sensor context tokens by applying
global average pooling over the initial sparse voxel features of each
modality after the input layer, and project them to the same dimension
via a shared linear layer:
$\mathbf{t}_{R},\, \mathbf{t}_{L} \in \mathbb{R}^{512}$.
These are concatenated with the condition token and fed into a
sensor-aware mixing MLP:
\begin{equation}
  \Delta\mathbf{c} = \mathrm{MLP}\!\left(
    [\,\mathbf{c} \;\|\; \mathbf{t}_{R} \;\|\; \mathbf{t}_{L}\,]
  \right) \in \mathbb{R}^{512},
  \label{eq:sensor_mlp}
\end{equation}
where $\|\cdot\|$ denotes concatenation and the MLP consists of two
linear layers with a hidden dimension of 1024.
The final condition token is obtained via a residual connection and
layer normalization,
$\hat{\mathbf{c}} = \mathrm{LN}(\mathbf{c} + \Delta\mathbf{c})
\in \mathbb{R}^{512}$,
and is used as the shared input to both the branch router and the
per-layer condition-gated fusion in the multi-branch backbone.
\subsection{Branch Routing over Multi-Branch Features}
\label{sec:branch_routing}

This module constructs three parallel sparse 3D feature streams from the
radar and LiDAR inputs, and uses the condition token $\hat{\mathbf{c}}$
to predict per-sample routing weights that adaptively combine them.

\textbf{Three parallel sparse 3D encoding streams.}
Both LiDAR and radar voxel features are processed through three-layer
sparse 3D convolutional encoders\cite{graham2015sparse}, where each layer $l \in \{1, 2, 3\}$
consists of a strided SparseConv3d for downsampling followed by two
SubMConv3d residual blocks, producing feature channels
$C_l \in \{64, 128, 256\}$.
We maintain three independent streams throughout the encoder:
a \textit{LiDAR-only branch} that processes LiDAR features
$\mathbf{L}^{l} \in \mathbb{R}^{N_l \times C_l}$ without any
radar interaction;
a \textit{Radar-only branch} that processes radar features
$\mathbf{R}^{l} \in \mathbb{R}^{M_l \times C_l}$ independently;
and a \textit{fusion branch} that starts from the same LiDAR features
as the LiDAR-only branch but receives radar-to-LiDAR cross-modal
augmentation at every layer, yielding fused features
$\mathbf{F}^{l} \in \mathbb{R}^{N_l \times C_l}$.
The LiDAR-only and fusion branches share the same convolutional weights,
so the only structural difference between them is the radar augmentation
applied within the fusion branch.

\textbf{Condition-gated KNN fusion.}
At each layer $l$ of the fusion branch, we augment each LiDAR voxel
feature using its spatially neighboring radar voxels.
For each LiDAR voxel $i$ with feature
$\mathbf{q}_i = \mathbf{L}^{l}_i \in \mathbb{R}^{C_l}$,
we retrieve its $K_l = \lfloor 64 / 2^{l-1} \rfloor$ nearest radar
voxels in 3D voxel index space and collect their features as keys
$\mathbf{K}_i \in \mathbb{R}^{K_l \times C_l}$.
A dot-product attention aggregates the radar context:
\begin{equation}
  \bar{\mathbf{a}}_i =
    \mathrm{softmax}\!\left(\mathbf{q}_i \mathbf{K}_i^{\top}\right)
    \mathbf{V}_i,
  \quad
  \mathbf{V}_i = \mathrm{Linear}(\mathbf{K}_i),
  \label{eq:knn_attn}
\end{equation}
where $\mathbf{V}_i \in \mathbb{R}^{K_l \times C_l}$ is a linear
projection of the keys.
The aggregated feature $\bar{\mathbf{a}}_i$ is then modulated by a
\textit{condition gate} derived from $\hat{\mathbf{c}}$.
Specifically, the condition token is projected to $\mathbb{R}^{C_l}$
via a layer-specific linear layer and concatenated with each radar key,
yielding gate logits that are globally average-pooled and passed through
a sigmoid activation:
\begin{equation}
  \mathbf{g}_i =
    \sigma\!\left(
      \mathrm{GAP}\!\left(
        \mathrm{Linear}([\,\mathbf{K}_i \;\|\; \hat{\mathbf{c}}^{l}\,])
      \right)
    \right) \in \mathbb{R}^{C_l},
  \label{eq:cond_gate}
\end{equation}
where $\hat{\mathbf{c}}^{l}$ is the layer-wise projected condition token.
The final fused feature at voxel $i$ is:
\begin{equation}
  \mathbf{F}^{l}_i = (\bar{\mathbf{a}}_i \odot \mathbf{g}_i) + \mathbf{q}_i,
  \label{eq:fusion_residual}
\end{equation}
where $\odot$ denotes element-wise multiplication and the $\mathbf{q}_i$
term forms a residual connection that preserves the original LiDAR geometry.
This gating mechanism allows the condition token to modulate how much
radar information is absorbed at each layer, depending on the current
weather context.

\textbf{BEV feature construction.}
After all three encoding layers, the 3D sparse features of each branch
are collapsed along the $z$-axis via a full-kernel SparseConv3d and
projected to the BEV plane via transposed convolution upsampling.
Per-layer BEV maps from all three layers are concatenated along the
channel dimension, producing branch-wise BEV feature maps
$\tilde{\mathbf{F}}_{L},\, \tilde{\mathbf{F}}_{R},\,
\tilde{\mathbf{F}}_{F} \in \mathbb{R}^{B \times 768 \times 32 \times 180}$.

\textbf{Weather-conditioned branch router.}
A two-layer MLP maps the condition token $\hat{\mathbf{c}}$ to a
three-dimensional routing logit vector, which is normalized via softmax
to obtain branch weights:
\begin{equation}
  \mathbf{w} = [w_L,\, w_R,\, w_F] =
    \mathrm{softmax}\!\left(\mathrm{MLP}(\hat{\mathbf{c}})\right).
  \label{eq:router}
\end{equation}
To prevent branch collapse during early training, we apply a weight floor
$\varepsilon = 0.1$ that guarantees a minimum contribution from each branch:
\begin{equation}
  w_i \;\leftarrow\; (1 - 3\varepsilon)\,w_i + \varepsilon,
  \quad \forall\, i \in \{L, R, F\}.
  \label{eq:weight_floor}
\end{equation}
The router is initialized with zero weights so that all three branches
start with equal contribution $w_L = w_R = w_F = 1/3$, and the
weights diverge as training progresses and the condition token acquires
discriminative weather information.

\textbf{Soft branch aggregation.}
The final BEV feature is assembled from the three branch-wise maps via
a structured soft weighting that preserves the physical semantics of
each branch:
\begin{equation}
  \tilde{\mathbf{F}} =
    \bigl[\,(w_R + w_F)\,\tilde{\mathbf{F}}_{R}
    \;\big\|\;
    w_L\,\tilde{\mathbf{F}}_{L} + w_F\,\tilde{\mathbf{F}}_{F}\,\bigr],
  \label{eq:soft_aggregation}
\end{equation}
where $\|\cdot\|$ denotes channel-wise concatenation.
The design reflects the underlying hard-selection logic of the three
branches: when $w_L \to 1$, the radar-side channel vanishes and only
pure LiDAR features are retained; when $w_R \to 1$, the lidar-side
channel vanishes and only radar features are active; when $w_F \to 1$,
the radar-augmented fusion features dominate.
Soft weighting thus enables a smooth, differentiable interpolation across
these three representational preferences as a function of the weather
condition.
$\tilde{\mathbf{F}} \in \mathbb{R}^{B \times 1536 \times 32 \times 180}$
is then passed to the detection head for final 3D object prediction.

\subsection{Training Objectives}
\label{sec:training}

Training the proposed framework involves four objectives that together
encourage both accurate detection and meaningful weather-dependent
routing behavior.

\textbf{Detection loss.}
The primary objective is the standard anchor-based detection loss
$\mathcal{L}_{\mathrm{det}}$, consisting of a focal classification loss over foreground and background anchors and a smooth-$\ell_1$ regression loss over the eight-dimensional box parameters $(x, y, z, w, l, h,$ 
$\sin\theta, \cos\theta)$ for matched positive anchors.

\textbf{Weather auxiliary classification loss.}
A key challenge in routing-based fusion is that the condition token
$\hat{\mathbf{c}}$ may not spontaneously retain discriminative weather
information when trained with the detection loss alone.
To address this, we attach a lightweight auxiliary head---a two-layer
MLP mapping $\hat{\mathbf{c}} \in \mathbb{R}^{512}$ to $K\!=\!7$
weather class logits---and supervise it with a class-weighted
cross-entropy loss using the ground-truth weather label $y \in \{0,\ldots,6\}$:
\begin{equation}
  \mathcal{L}_{\mathrm{aux}} =
    -\rho_{y} \log \frac{e^{z_y}}{\sum_{k} e^{z_k}},
  \label{eq:aux_loss}
\end{equation}
where $z_k$ are the predicted logits and $\rho_k$ is the class weight
for weather condition $k$.
Since adverse weather conditions such as sleet, heavy snow, and fog are
underrepresented in the dataset, we assign higher weights
$\rho \in [1.0,\, 2.0]$ to these rare classes to prevent the token
from collapsing toward the dominant normal/overcast distribution.
This auxiliary objective ensures that $\hat{\mathbf{c}}$ preserves
discriminative weather information and thus provides a meaningful
basis for routing.

\textbf{Weather diversity regularization.}
Even with a well-trained condition token, the branch router may learn
to assign similar routing weights regardless of the weather category.
To explicitly encourage weather-dependent routing patterns, we introduce
a diversity regularization term $\mathcal{L}_{\mathrm{div}}$ that
operates on the routing weight vectors $\mathbf{w} \in \mathbb{R}^3$
of all samples within a mini-batch.
For each weather class $k$ present in the batch, we compute the mean
routing vector $\boldsymbol{\mu}_k$ and penalize intra-class variance
to promote consistent routing within the same condition:
\begin{equation}
  \mathcal{L}_{\mathrm{intra}} =
    \frac{1}{|\mathcal{K}|} \sum_{k \in \mathcal{K}}
    \frac{1}{|\mathcal{B}_k|} \sum_{i \in \mathcal{B}_k}
    \|\mathbf{w}_i - \boldsymbol{\mu}_k\|^2,
  \label{eq:intra_loss}
\end{equation}
where $\mathcal{K}$ is the set of unique weather classes in the batch
and $\mathcal{B}_k$ the corresponding sample indices.
We further enforce inter-class separation by penalizing pairs of class
centers that are too close in routing space via a hinge loss:
\begin{equation}
  \mathcal{L}_{\mathrm{inter}} =
    \frac{1}{\binom{|\mathcal{K}|}{2}}
    \sum_{j < k}
    \max\!\left(0,\; m - \|\boldsymbol{\mu}_j - \boldsymbol{\mu}_k\|_2\right),
  \label{eq:inter_loss}
\end{equation}
where $m\!=\!0.12$ is the margin.
The combined diversity loss is
$\mathcal{L}_{\mathrm{div}} =
\mathcal{L}_{\mathrm{intra}} + \mathcal{L}_{\mathrm{inter}}$.

\textbf{Branch entropy regularization.}
To prevent the router from collapsing to a near-deterministic single
branch early in training, we introduce an entropy-based regularization
term.
The normalized routing entropy for a batch is defined as:
\begin{equation}
  \bar{H} =
    \frac{1}{B} \sum_{i=1}^{B}
    \frac{-\sum_{j} w_{ij} \log w_{ij}}{\log 3},
  \label{eq:entropy}
\end{equation}
where $\log 3$ normalizes the entropy to the range $[0, 1]$.
We penalize configurations where the routing is overly concentrated
(i.e., when $\bar{H}$ falls below a target ratio $\tau$):
\begin{equation}
  \mathcal{L}_{\mathrm{ent}} = \max(0,\; \tau - \bar{H}),
  \label{eq:entropy_loss}
\end{equation}
with $\tau\!=\!0.78$, which corresponds to routing distributions that
are moderately spread across the three branches rather than dominated
by a single one.
Together with the weight floor mechanism in Eq.~\eqref{eq:weight_floor},
this ensures that all branches remain active and meaningful throughout
training.

\textbf{Total loss and training strategy.}
The full training objective is:
\begin{equation}
  \mathcal{L} =
    \mathcal{L}_{\mathrm{det}}
    + \lambda_{\mathrm{aux}}\,\mathcal{L}_{\mathrm{aux}}
    + \lambda_{\mathrm{div}}\,\mathcal{L}_{\mathrm{div}}
    + \lambda_{\mathrm{ent}}\,\mathcal{L}_{\mathrm{ent}},
  \label{eq:total_loss}
\end{equation}
with $\lambda_{\mathrm{aux}}\!=\!0.1$,
$\lambda_{\mathrm{div}}\!=\!0.02$, and
$\lambda_{\mathrm{ent}}\!=\!0.01$.
Note that all weather labels are used only during training as
supervisory signals for $\mathcal{L}_{\mathrm{aux}}$ and
$\mathcal{L}_{\mathrm{div}}$; at inference, no weather label is
required and routing is performed solely from the condition token.


\section{Experiments}

\begin{table*}[t]
\centering
\caption{Quantitative results of LiDAR and 4D radar-based 3D object detection methods on K-Radar dataset. We present the modality of each method (L: LiDAR, 4DR: 4D radar) and detailed performance for each weather condition. Best in bold with blue background (\best{best}), second in light blue background (\second{second}). * denotes our reproduced results.}
\label{tab:kradar_table1}
\resizebox{\textwidth}{!}{
\begin{tabular}{l|c|cccccccccc}
\toprule
\multirow{2}{*}{Methods} & \multirow{2}{*}{Modality} & \multirow{2}{*}{IoU} & \multirow{2}{*}{Metric} & \multicolumn{8}{c}{Weather Condition} \\
\cmidrule(lr){5-12}
& & & & Total & Normal & Overcast & Fog & Rain & Sleet & Lightsnow & Heavysnow \\
\midrule
\multirow{4}{*}{RTNH~\cite{paek2022kradar}} 
& \multirow{4}{*}{4DR} 
& \multirow{2}{*}{0.3} 
& $\textit{AP}_{\mathrm{BEV}}$ & 41.1 & 41.0 & 44.6 & 45.4 & 32.9 & 50.6 & 81.5 & 56.3 \\
&  &  & $\textit{AP}_{\mathrm{3D}}$ & 37.4 & 37.6 & 42.0 & 41.2 & 29.2 & 49.1 & 63.9 & 43.1 \\
\cmidrule(lr){3-12}
&  & \multirow{2}{*}{0.5} 
& $\textit{AP}_{\mathrm{BEV}}$ & 36.0 & 35.8 & 41.9 & 44.8 & 30.2 & 34.5 & 63.9 & \second{55.1} \\
&  &  & $\textit{AP}_{\mathrm{3D}}$ & 14.1 & 19.7 & 20.5 & 15.9 & 13.0 & 13.5 & 21.0 & 6.36 \\
\midrule
\multirow{4}{*}{RTNH*~\cite{paek2022kradar}} 
& \multirow{4}{*}{L} 
& \multirow{2}{*}{0.3} 
& $\textit{AP}_{\mathrm{BEV}}$ & 76.5 & 76.5 & 88.2 & 86.3 & 77.3 & \second{55.3} & 81.1 & 59.5 \\
&  &  & $\textit{AP}_{\mathrm{3D}}$ & \second{72.7} & \second{73.1} & 76.5 & 84.8 & 64.5 & \best{53.4} & \second{80.3} & 52.9 \\
\cmidrule(lr){3-12}
&  & \multirow{2}{*}{0.5} 
& $\textit{AP}_{\mathrm{BEV}}$ & \second{66.3} & \second{65.4} & \second{87.4} & \second{83.8} & \second{73.7} & \second{48.8} & \second{78.5} & 48.1 \\
&  &  & $\textit{AP}_{\mathrm{3D}}$ & \second{37.8} & \second{39.8} & \second{46.3} & \second{59.8} & 28.2 & \best{31.4} & 50.7 & 24.6 \\
\midrule
\multirow{4}{*}{PointPillars~\cite{lang2018pointpillars}} 
& \multirow{4}{*}{L} 
& \multirow{2}{*}{0.3} 
& $\textit{AP}_{\mathrm{BEV}}$ & 51.9 & 51.6 & 53.5 & 45.4 & 44.7 & 54.3 & 81.2 & 55.2 \\
&  &  & $\textit{AP}_{\mathrm{3D}}$ & 47.3 & 46.7 & 51.9 & 44.8 & 42.4 & 45.5 & 59.2 & \second{55.2} \\
\cmidrule(lr){3-12}
&  & \multirow{2}{*}{0.5} 
& $\textit{AP}_{\mathrm{BEV}}$ & 49.1 & 48.2 & 53.0 & 45.4 & 44.2 & 45.9 & 74.5 & 53.8 \\
&  &  & $\textit{AP}_{\mathrm{3D}}$ & 22.4 & 21.8 & 28.0 & 28.2 & 27.2 & 22.6 & 23.2 & 12.9 \\
\midrule
\multirow{4}{*}{InterFusion~\cite{wang2022interfusion}} 
& \multirow{4}{*}{4DR+L} 
& \multirow{2}{*}{0.3} 
& $\textit{AP}_{\mathrm{BEV}}$ & 57.5 & 57.2 & 60.8 & 81.2 & 52.8 & 27.5 & 72.6 & 57.2 \\
&  &  & $\textit{AP}_{\mathrm{3D}}$ & 53.0 & 51.1 & 58.1 & 80.9 & 40.4 & 23.0 & 71.0 & \second{55.2} \\
\cmidrule(lr){3-12}
&  & \multirow{2}{*}{0.5} 
& $\textit{AP}_{\mathrm{BEV}}$ & 52.9 & 50.0 & 59.0 & 80.3 & 50.0 & 22.7 & 72.2 & 53.3 \\
&  &  & $\textit{AP}_{\mathrm{3D}}$ & 17.5 & 15.3 & 20.5 & 47.6 & 12.9 & 9.33 & \second{56.8} & \second{25.7} \\
\midrule
\multirow{4}{*}{3D-LRF*~\cite{chae20243dlrf}} 
& \multirow{4}{*}{4DR+L} 
& \multirow{2}{*}{0.3} 
& $\textit{AP}_{\mathrm{BEV}}$ & \second{80.0} & \second{81.2} & \second{89.0} & \second{88.8} & \best{82.6} & 42.9 & \second{85.6} & \second{61.5} \\
&  &  & $\textit{AP}_{\mathrm{3D}}$ & 70.4 & 70.9 & \second{84.1} & \second{86.9} & \second{71.1} & 36.8 & 79.5 & 55.1 \\
\cmidrule(lr){3-12}
&  & \multirow{2}{*}{0.5} 
& $\textit{AP}_{\mathrm{BEV}}$ & 64.1 & 64.7 & 78.7 & 77.8 & 71.4 & 36.2 & \second{78.5} & 45.7 \\
&  &  & $\textit{AP}_{\mathrm{3D}}$ & 30.0 & 31.2 & 40.0 & 32.3 & \second{28.5} & 21.4 & 52.8 & 24.5 \\
\midrule
\multirow{4}{*}{Ours} 
& \multirow{4}{*}{4DR+L} 
& \multirow{2}{*}{0.3} 
& $\textit{AP}_{\mathrm{BEV}}$ & \best{81.2} & \best{82.1} & \best{92.5} & \best{89.9} & \second{82.3} & \best{58.9} & \best{88.5} & \best{67.9} \\
&  &  & $\textit{AP}_{\mathrm{3D}}$ & \best{76.8} & \best{77.3} & \best{88.2} & \best{87.8} & \best{77.8} & \second{51.0} & \best{86.6} & \best{61.4} \\
\cmidrule(lr){3-12}
&  & \multirow{2}{*}{0.5} 
& $\textit{AP}_{\mathrm{BEV}}$ & \best{71.5} & \best{70.1} & \best{88.3} & \best{88.1} & \best{78.0} & \best{49.0} & \best{83.0} & \best{57.2} \\
&  &  & $\textit{AP}_{\mathrm{3D}}$ & \best{43.5} & \best{41.5} & \best{63.3} & \best{63.4} & \best{42.2} & \second{30.5} & \best{66.8} & \best{33.7} \\
\bottomrule
\end{tabular}
}
\end{table*}

\subsection{Experimental Setup}
\textbf{Dataset and Benchmark.} We evaluate our method on the K-Radar dataset\cite{paek2022kradar}, following the same experimental protocol as 3D-LRF\cite{chae20243dlrf} for fair comparison. K-Radar is a large-scale autonomous driving benchmark that provides synchronized LiDAR, 4D radar, and camera data under diverse weather conditions, including normal, rain, fog, snow, sleet, and overcast scenes. Following prior work, we conduct evaluation on the \textit{Sedan} category and adopt the official benchmark metrics, namely AP$_{3D}$ and AP$_{BEV}$. 

\textbf{Implementation Details}
All models are trained on the K-Radar dataset using the official train/test split.
Input point clouds and radar sparse cubes are cropped to a driving-corridor ROI of
$x \in [0, 72]$\,m, $y \in [-6.4, 6.4]$\,m, $z \in [-2, 6]$\,m with a voxel grid size of 0.4\,m.
The backbone encodes features at three levels with channel dimensions $\{64, 128, 256\}$.
Sedan anchors of size $4.2 \times 2.1 \times 2.0$\,m are placed at rotations $\{0, \pi/2\}$,
with positive/negative IoU thresholds of 0.5/0.2.
The branch weight floor is set to $\varepsilon = 0.1$, the diversity margin to $m = 0.12$,
and the entropy target to $\tau = 0.78$.
The loss weights are $\lambda_{\text{aux}} = 0.1$, $\lambda_{\text{div}} = 0.02$,
and $\lambda_{\text{ent}} = 0.01$.
The network is optimized with Adam ($\beta_{1}=0.9$, $\beta_{2}=0.999$,
weight decay $= 0.01$) at an initial learning rate of $5\times10^{-4}$,
decayed to $1\times10^{-4}$ via cosine annealing over 20 epochs with a batch size of 4.

\subsection{Comparison with SOTA Methods}
Compared with prior methods in Table~1, our approach establishes a new state of the art on K-Radar, with the most significant gains achieved over the previous strongest fusion baseline, 3D-LRF*\cite{chae20243dlrf}. While LiDAR-only, radar-only, and earlier fusion methods broadly demonstrate the advantage of multi-modal perception, the comparison with 3D-LRF*\cite{chae20243dlrf} is the most critical since it represents the most competitive prior result. Specifically, under IoU$=0.3$, our method achieves an improvement of $+1.2$ in AP$_{\mathrm{BEV}}$ (from 80.0 to 81.2) and $+6.4$ in AP$_{\mathrm{3D}}$ (from 70.4 to 76.8). Under the more stringent IoU$=0.5$ setting, the improvements become substantially larger, with AP$_{\mathrm{BEV}}$ increasing by $+7.4$ (from 64.1 to 71.5) and AP$_{\mathrm{3D}}$ increasing by $+13.5$ (from 30.0 to 43.5). Such results indicate that the proposed weather-conditioned branch routing is particularly effective for precise 3D localization, where fixed fusion strategies tend to degrade more severely. Notably, the superiority of our method is also consistent across adverse-weather scenarios. Compared with 3D-LRF*\cite{chae20243dlrf}, our method achieves clear gains in overcast conditions, improving AP$_{\mathrm{3D}}$ from 84.1 to 88.2 at IoU$=0.3$ and from 40.0 to 63.3 at IoU$=0.5$, while AP$_{\mathrm{BEV}}$ increases from 89.0 to 92.5 and from 78.7 to 88.3, respectively. In foggy scenes, our method further improves AP$_{\mathrm{BEV}}$/AP$_{\mathrm{3D}}$ from 88.8/86.9 to 89.9/87.8 at IoU$=0.3$, and more substantially from 77.8/32.3 to 88.1/63.4 at IoU$=0.5$. Similar trends are observed in snow conditions: under light snow, AP$_{\mathrm{3D}}$ is improved from 79.5 to 86.6 at IoU$=0.3$ and from 52.8 to 66.8 at IoU$=0.5$, while under heavy snow, AP$_{\mathrm{BEV}}$/AP$_{\mathrm{3D}}$ rises from 61.5/55.1 to 67.9/61.4 at IoU$=0.3$ and from 45.7/24.5 to 57.2/33.7 at IoU$=0.5$. Overall, these results suggest that explicitly modeling weather-dependent preference over LiDAR-only, radar-only, and fused representations is more effective than relying on a fixed fusion pathway, leading to stronger robustness and better generalization across diverse adverse-weather conditions.

\begin{table*}[t]
\centering
\caption{Ablation study of the proposed method on K-Radar. Branch routing denotes the weather-conditioned multi-branch aggregation, $L_{\mathrm{aux}}$ is the auxiliary weather classification loss, and $L_{\mathrm{div}}$ is the weather diversity regularization.}
\label{tab:ablation}
\resizebox{0.92\textwidth}{!}{
\begin{tabular}{l|ccc|cccc}
\toprule
Method & Branch Routing & $L_{\mathrm{aux}}$ & $L_{\mathrm{div}}$ & 
AP$_{\mathrm{BEV}}^{0.5}$ & AP$_{\mathrm{3D}}^{0.5}$ & AP$_{\mathrm{BEV}}^{0.3}$ & AP$_{\mathrm{3D}}^{0.3}$ \\
\midrule
w/o branch routing (force Fusion only) &  & \checkmark & \checkmark & 54.8 & 22.0 & 78.6 & 68.8 \\
w/o weather supervision ($L_{\mathrm{aux}}$, $L_{\mathrm{div}}$) & \checkmark &  &  & 70.7 & 43.0 & 77.4 & 73.9 \\
w/o $L_{\mathrm{div}}$ & \checkmark & \checkmark &  & 64.8 & 30.2 & 78.3 & 72.9 \\
Full model & \checkmark & \checkmark & \checkmark & \textbf{71.5} & \textbf{43.5} & \textbf{81.2} & \textbf{76.8} \\
\bottomrule
\end{tabular}
}
\end{table*}
\begin{table*}[t]
\centering

\caption{Comparison of different routing designs on K-Radar. The MLP-based router (Ours) achieves the best overall performance and shows stronger robustness across adverse weather conditions.}
\vspace{-0.3em}
\label{tab:routing_design}
\resizebox{\textwidth}{!}{
\begin{tabular}{l|cc|cc|cc|cc|cc|cc}
\toprule
\multirow{2}{*}{Method} 
& \multicolumn{2}{c|}{All} 
& \multicolumn{2}{c|}{Overcast} 
& \multicolumn{2}{c|}{Fog} 
& \multicolumn{2}{c|}{Sleet} 
& \multicolumn{2}{c|}{Light Snow} 
& \multicolumn{2}{c}{Heavy Snow} \\
& BEV@0.5 & 3D@0.5 & BEV@0.5 & 3D@0.5 & BEV@0.5 & 3D@0.5 & BEV@0.5 & 3D@0.5 & BEV@0.5 & 3D@0.5 & BEV@0.5 & 3D@0.5 \\
\midrule
Transformer router & 48.17 & 27.31 & 30.83 & 7.40 & 63.74 & 48.17 & 40.77 & 27.66 & 65.72 & 43.69 & 53.30 & 35.33 \\
Cross-attention router & 56.8 & 33.9 & 41.2 & 18.6 & 71.5 & 53.8 & 44.8 & 28.7 & 71.4 & 49.6 & 54.9 & 28.1 \\
MLP router (ours) & \textbf{71.5} & \textbf{43.5} & \textbf{88.3} & \textbf{63.3} & \textbf{88.1} & \textbf{63.4} & \textbf{49.0} & \textbf{30.5} & \textbf{83.0} & \textbf{66.8} & \textbf{57.2} & \textbf{33.7} \\
\bottomrule
\end{tabular}
}
\end{table*}

\subsection{Ablation Study}
\textbf{Effect of Each Component.}
Tab.~\ref{tab:ablation} presents the ablation study of each component. First, removing branch routing and forcing the model to use only the fusion branch leads to a clear performance drop, reducing AP$_{\mathrm{BEV}}$/AP$_{\mathrm{3D}}$ at IoU$=0.5$ from 71.5/43.5 to 54.8/22.0. This substantial degradation indicates that the main gain of our method does not come from simply introducing an additional fusion pathway, but from explicitly enabling weather-conditioned selection among LiDAR-only, radar-only, and fused representations. Second, removing the weather supervision terms ($L_{\mathrm{aux}}$ and $L_{\mathrm{div}}$) causes the performance to decrease to 70.7 AP$_{\mathrm{BEV}}$ and 43.0 AP$_{\mathrm{3D}}$ at IoU$=0.5$, and to 77.4/73.9 at IoU$=0.3$, showing that weather-aware supervision helps the router learn more discriminative condition-dependent preferences. Third, removing only the diversity regularization results in a much larger drop, especially under the stricter IoU$=0.5$ setting, where AP$_{\mathrm{BEV}}$ and AP$_{\mathrm{3D}}$ fall to 64.8 and 30.2, respectively. This suggests that $L_{\mathrm{div}}$ plays a critical role in preventing the router from collapsing to similar branch preferences across weather conditions. The effect is particularly evident in adverse weather: for example, without branch routing, the AP$_{\mathrm{3D}}$ at IoU$=0.5$ drops to 11.1 in fog and 5.3 in heavy snow, while removing $L_{\mathrm{div}}$ reduces the corresponding performance to 40.2 and 9.8, both far below the full model (63.4 and 33.7). Overall, these results verify that both the routing mechanism and the weather-aware learning objectives are necessary, and that the diversity regularization is especially important for achieving robust performance across challenging weather conditions.

\textbf{Effect of Routing Module Design.}
We further compare different routing designs, including a Transformer-based router, a cross-attention router, and our MLP-based router. As shown in Tab.~\ref{tab:routing_design}, the proposed MLP router achieves the best overall performance, reaching 71.5 AP$_{\mathrm{BEV}}$ and 43.5 AP$_{\mathrm{3D}}$ at IoU@0.5, which is substantially higher than the Transformer-based design (48.17/27.31) and also better than the cross-attention variant. Specifically, compared with the Transformer router, the MLP design achieves an improvement of +23.3 AP$_{\mathrm{BEV}}$ and +16.2 AP$_{\mathrm{3D}}$ on the full set. The advantage is even more evident in adverse weather conditions. In overcast scenes, the MLP router improves AP$_{\mathrm{BEV}}$/AP$_{\mathrm{3D}}$ from 30.83/7.40 to 88.3/63.3; in fog, it further increases the performance from 63.74/48.17 to 88.1/63.4. Similar gains are observed under snow conditions, where the MLP router achieves 83.0/66.8 in light snow and 57.2/33.7 in heavy snow, consistently outperforming both alternative designs. Although the cross-attention router provides a stronger baseline than the Transformer router, it still remains inferior to the MLP design, suggesting that a lightweight and direct mapping from the weather-conditioned token to branch weights is more effective than more complicated token-interaction mechanisms. Overall, these results show that MLP-based router offers a better trade-off between adaptivity, optimization stability, and detection performance.

\textbf{Effect of Hyperparameters.}
We further conduct an ablation study to investigate the impact of key hyperparameters in our training objectives: the auxiliary weather classification weight $\lambda_{aux}$, the weather diversity regularization weight $\lambda_{div}$, and the branch entropy regularization weight $\lambda_{ent}$. As shown in Table \ref{tab:key_hparams}, the model achieves its best performance with the default setting ($\lambda_{aux}=0.10$, $\lambda_{div}=0.02$, $\lambda_{ent}=0.010$). Notably, the diversity weight $\lambda_{div}$ plays the most critical role. When $\lambda_{div}$ is set to 0.00, the performance suffers a drastic drop (from 43.5\% to 30.2\% in $AP_{3D}$), verifying that without explicit diversity penalties, the router fails to maintain distinguishable branch preferences across different weather conditions and collapses into a degenerate state. Furthermore, the entropy regularization $\lambda_{ent}$ is shown to be necessary to prevent deterministic branch collapse during early training stages (yielding a drop to 35.9\% $AP_{3D}$ when set to 0.000). However, an excessively large value ($\lambda_{ent}=0.020$) forces the routing distribution to be overly uniform, which hinders the model's ability to adaptively select the dominant modality and thus degrades performance. Finally, varying the auxiliary weather supervision weight $\lambda_{aux}$ demonstrates that explicitly guiding the condition token provides a stable performance boost, with $\lambda_{aux}=0.10$ striking the optimal balance between weather-aware representation learning and the primary detection task.

\begin{table}[t]
\centering
\caption{Ablation study on key hyperparameters of the proposed method. We evaluate the impact of the auxiliary weather supervision weight ($\lambda_{\text{aux}}$), diversity regularization weight ($\lambda_{\text{div}}$), and entropy regularization weight ($\lambda_{\text{ent}}$). Best results are highlighted in bold.}
\label{tab:key_hparams}
\setlength{\tabcolsep}{8pt}
\renewcommand{\arraystretch}{1.15}
\resizebox{0.46\textwidth}{!}{
\begin{tabular}{l|ccc}
\hline
Hyperparameter & Value & AP$_{\mathrm{3D}}^{0.5}$ (\%) & AP$_{\mathrm{BEV}}^{0.5}$ (\%) \\
\hline
\multirow{4}{*}{$\lambda_{\text{aux}}$} & 0.00 & 27.5 & 56.3 \\
                       & 0.05 & 43.5 & 70.8 \\
                       & \textbf{0.10} & \textbf{43.5} & \textbf{71.5} \\
                       & 0.20 & 33.9 & 63.4 \\
\hline
\multirow{4}{*}{$\lambda_{\text{div}}$} & 0.00 & 30.2 & 64.8 \\
                       & 0.01 & 37.9 & 68.7 \\
                       & \textbf{0.02} & \textbf{43.5} & \textbf{71.5} \\
                       & 0.05 & 41.5 & 70.1 \\
\hline
\multirow{4}{*}{$\lambda_{\text{ent}}$} & 0.000 & 35.9 & 63.6 \\
                       & 0.005 & 42.6 & 70.8 \\
                       & \textbf{0.010} & \textbf{43.5} & \textbf{71.5} \\
                       & 0.020 & 42.1 & 70.4 \\
\hline
\end{tabular}}
\vspace{-1.5em}
\end{table}

\begin{figure}[!ht]
    \centering
    \includegraphics[width=0.48\textwidth]{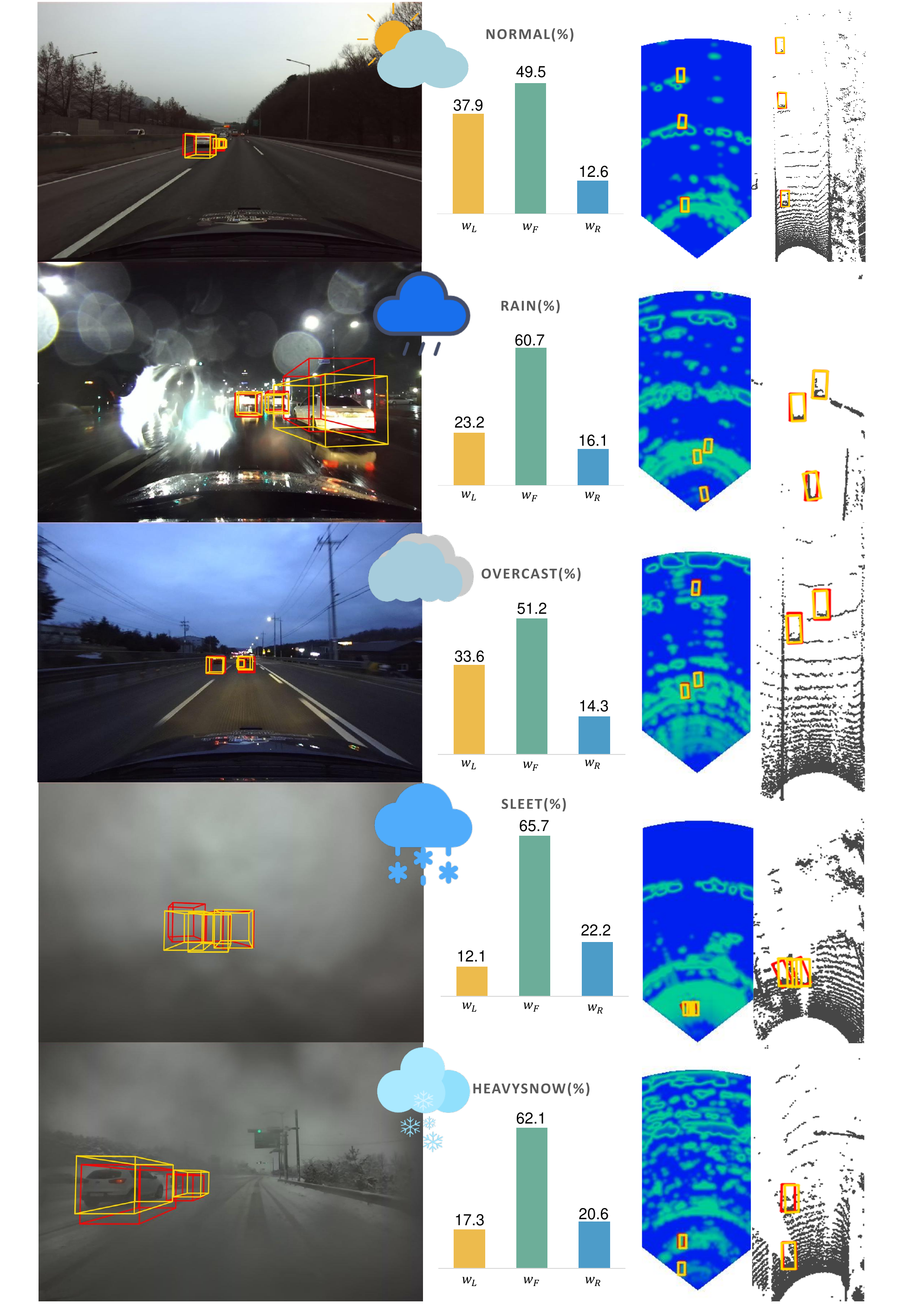}
    \caption{Qualitative results and interpretable routing behavior under diverse weather conditions. From top to bottom: Normal, Rain, Overcast, Sleet and Heavy Snow scenarios. 
    For each scene, we visualize the camera view with 3D bounding box predictions (left), the adaptive branch weights $w_L, w_F, w_R$ predicted by the weather-conditioned router (middle), and the corresponding radar and LiDAR BEV representations (right). 
    }
    \Description{}
    \label{fig:qualitative}

\end{figure}

\subsection{Qualitative Analysis}
Fig. 3 visualizes the 3D detection results alongside the dynamic routing weights ($w_L$, $w_F$, $w_R$) under Normal, Rain, and Heavy Snow conditions, demonstrating our model's interpretable modality preference. In normal weather (top row), the router naturally favors the accurate geometry of the pure LiDAR ($w_L=37.9\%$) and fusion ($w_F=49.5\%$) branches.  However, as weather deteriorates and LiDAR signals suffer from scattering, the model adaptively shifts its reliance toward the more robust 4D radar. In rainy conditions (middle row), the fusion weight $w_F$ increases to 60.7\%. Under severe heavy snow (bottom row), where LiDAR point clouds are heavily corrupted, the pure LiDAR branch is explicitly suppressed ($w_L$ drops to 17.3\%), while the network relies predominantly on the radar-augmented fusion ($w_F=62.1\%$) and pure radar ($w_R=20.6\%$) branches. This logical adaptation aligns perfectly with sensor physics, proving that our framework ensures high detection accuracy while providing a transparent, condition-aware fusion process. By continuously recalibrating these weights, the network effectively mitigates the risk of severe performance degradation that typically plagues fixed-fusion detectors in harsh environments. Furthermore, this dynamic balancing acts as a built-in diagnostic tool, allowing developers to visually verify that the model shifts its sensor reliance safely and predictably. Furthermore, this confirms that our routing mechanism successfully avoids modality collapse, maintaining active utilization of all available branches.


\section{Conclusion}
In this paper, we propose a novel weather-conditioned branch routing framework for robust LiDAR-Radar 3D object detection. Diverging from conventional fixed fusion pipelines, our method dynamically aggregates LiDAR-only, radar-only, and fused representations based on the current environmental context. By incorporating weather-supervised learning objectives—specifically auxiliary weather classification and diversity regularization—we effectively stabilize the routing process and prevent branch collapse during training. Extensive evaluations on K-Radar benchmark demonstrate that our approach establishes a new SOTA performance across diverse challenging conditions. By decoupling the candidate representations from the automated routing decision, our architecture preserves analytical transparency without sacrificing precise localization accuracy. 
This structural advantage ensures that the system is uniquely suited for safety-critical deployment, where understanding the reasoning behind modality shifts is fundamentally just as critical as the raw detection output. 
More importantly, the proposed explicit routing mechanism offers highly interpretable insights into weather-aware modality preferences, highlighting a practical and transparent alternative for robust multi-sensor perception.
Ultimately, we believe this adaptive paradigm paves the way for safer, more reliable autonomous driving systems capable of navigating unpredictable, real-world environments.


\bibliographystyle{ACM-Reference-Format}
\bibliography{ref}

@String{Computing = "Computing" }

@String{Computer = "{IEEE} Computer" }

@String{Springer = "Springer-Verlag" }

@ArtifactSoftware{R,
    title = {R: A Language and Environment for Statistical Computing},
    author = {{R Core Team}},
    organization = {R Foundation for Statistical Computing},
    address = {Vienna, Austria},
    year = {2019},
    url = {https://www.R-project.org/},
}

@article{paek2022kradar,
  title={K-radar: 4d radar object detection for autonomous driving in various weather conditions},
  author={Paek, Dong-Hee and Kong, Seung-Hyun and Wijaya, Kevin Tirta},
  journal={Advances in Neural Information Processing Systems},
  volume={35},
  pages={3819--3829},
  year={2022}
}

@inproceedings{lang2018pointpillars,
  title={Pointpillars: Fast encoders for object detection from point clouds},
  author={Lang, Alex H and Vora, Sourabh and Caesar, Holger and Zhou, Lubing and Yang, Jiong and Beijbom, Oscar},
  booktitle={Proceedings of the IEEE/CVF conference on computer vision and pattern recognition},
  pages={12697--12705},
  year={2019}
}

@inproceedings{wang2022interfusion,
  title={InterFusion: Interaction-based 4D radar and LiDAR fusion for 3D object detection},
  author={Wang, Li and Zhang, Xinyu and Xv, Baowei and Zhang, Jinzhao and Fu, Rong and Wang, Xiaoyu and Zhu, Lei and Ren, Haibing and Lu, Pingping and Li, Jun and others},
  booktitle={2022 IEEE/RSJ International Conference on Intelligent Robots and Systems (IROS)},
  pages={12247--12253},
  year={2022},
  organization={IEEE}
}

@inproceedings{chae20243dlrf,
  title={Towards robust 3d object detection with lidar and 4d radar fusion in various weather conditions},
  author={Chae, Yujeong and Kim, Hyeonseong and Yoon, Kuk-Jin},
  booktitle={Proceedings of the IEEE/CVF Conference on Computer Vision and Pattern Recognition},
  pages={15162--15172},
  year={2024}
}

@inproceedings{Bijelic2020seeing,
  title={Seeing through fog without seeing fog: Deep multimodal sensor fusion in unseen adverse weather},
  author={Bijelic, Mario and Gruber, Tobias and Mannan, Fahim and Kraus, Florian and Ritter, Werner and Dietmayer, Klaus and Heide, Felix},
  booktitle={Proceedings of the IEEE/CVF conference on computer vision and pattern recognition},
  pages={11682--11692},
  year={2020}
}

@article{song2024robustness,
  title={Robustness-aware 3d object detection in autonomous driving: A review and outlook},
  author={Song, Ziying and Liu, Lin and Jia, Feiyang and Luo, Yadan and Jia, Caiyan and Zhang, Guoxin and Yang, Lei and Wang, Li},
  journal={IEEE Transactions on Intelligent Transportation Systems},
  volume={25},
  number={11},
  pages={15407--15436},
  year={2024},
  publisher={IEEE}
}

@inproceedings{huang2025l4dr,
  title={L4dr: Lidar-4dradar fusion for weather-robust 3d object detection},
  author={Huang, Xun and Xu, Ziyu and Wu, Hai and Wang, Jinlong and Xia, Qiming and Xia, Yan and Li, Jonathan and Gao, Kyle and Wen, Chenglu and Wang, Cheng},
  booktitle={Proceedings of the AAAI conference on artificial intelligence},
  volume={39},
  number={4},
  pages={3806--3814},
  year={2025}
}

@inproceedings{liu2023bevfusion,
  title={Bevfusion: Multi-task multi-sensor fusion with unified bird's-eye view representation},
  author={Liu, Zhijian and Tang, Haotian and Amini, Alexander and Yang, Xinyu and Mao, Huizi and Rus, Daniela L and Han, Song},
  booktitle={2023 IEEE international conference on robotics and automation (ICRA)},
  pages={2774--2781},
  year={2023},
  organization={IEEE}
}

@inproceedings{bai2022transfusion,
  title={Transfusion: Robust lidar-camera fusion for 3d object detection with transformers},
  author={Bai, Xuyang and Hu, Zeyu and Zhu, Xinge and Huang, Qingqiu and Chen, Yilun and Fu, Hongbo and Tai, Chiew-Lan},
  booktitle={Proceedings of the IEEE/CVF conference on computer vision and pattern recognition},
  pages={1090--1099},
  year={2022}
}

@inproceedings{li2022deepfusion,
  title={Deepfusion: Lidar-camera deep fusion for multi-modal 3d object detection},
  author={Li, Yingwei and Yu, Adams Wei and Meng, Tianjian and Caine, Ben and Ngiam, Jiquan and Peng, Daiyi and Shen, Junyang and Lu, Yifeng and Zhou, Denny and Le, Quoc V and others},
  booktitle={Proceedings of the IEEE/CVF conference on computer vision and pattern recognition},
  pages={17182--17191},
  year={2022}
}

@inproceedings{chen2023futr3d,
  title={Futr3d: A unified sensor fusion framework for 3d detection},
  author={Chen, Xuanyao and Zhang, Tianyuan and Wang, Yue and Wang, Yilun and Zhao, Hang},
  booktitle={proceedings of the IEEE/CVF conference on computer vision and pattern recognition},
  pages={172--181},
  year={2023}
}

@inproceedings{zhao2025unibevfusion,
  title={Unibevfusion: Unified radar-vision bevfusion for 3d object detection},
  author={Zhao, Haocheng and Guan, Runwei and Wu, Taoyu and Man, Ka Lok and Yu, Limin and Yue, Yutao},
  booktitle={2025 IEEE International Conference on Robotics and Automation (ICRA)},
  pages={6321--6327},
  year={2025},
  organization={IEEE}
}

@inproceedings{yin2024fusion,
  title={Is-fusion: Instance-scene collaborative fusion for multimodal 3d object detection},
  author={Yin, Junbo and Shen, Jianbing and Chen, Runnan and Li, Wei and Yang, Ruigang and Frossard, Pascal and Wang, Wenguan},
  booktitle={Proceedings of the IEEE/CVF conference on computer vision and pattern recognition},
  pages={14905--14915},
  year={2024}
}

@inproceedings{huang2024sunshine,
  title={Sunshine to rainstorm: Cross-weather knowledge distillation for robust 3d object detection},
  author={Huang, Xun and Wu, Hai and Li, Xin and Fan, Xiaoliang and Wen, Chenglu and Wang, Cheng},
  booktitle={Proceedings of the AAAI Conference on Artificial Intelligence},
  volume={38},
  number={3},
  pages={2409--2416},
  year={2024}
}

@article{li2025moe3d,
  title={MoE3D: Mixture of Experts meets Multi-Modal 3D Understanding},
  author={Li, Yu and Hou, Yuenan and Wei, Yingmei and Zhu, Xinge and Ma, Yuexin and Shao, Wenqi and Guo, Yanming},
  journal={arXiv preprint arXiv:2511.22103},
  year={2025}
}

@inproceedings{sindagi2019mvx,
  title={Mvx-net: Multimodal voxelnet for 3d object detection},
  author={Sindagi, Vishwanath A and Zhou, Yin and Tuzel, Oncel},
  booktitle={2019 International Conference on Robotics and Automation (ICRA)},
  pages={7276--7282},
  year={2019},
  organization={IEEE}
}

@inproceedings{vora2020pointpainting,
  title={Pointpainting: Sequential fusion for 3d object detection},
  author={Vora, Sourabh and Lang, Alex H and Helou, Bassam and Beijbom, Oscar},
  booktitle={Proceedings of the IEEE/CVF conference on computer vision and pattern recognition},
  pages={4604--4612},
  year={2020}
}

@inproceedings{huang2020epnet,
  title={Epnet: Enhancing point features with image semantics for 3d object detection},
  author={Huang, Tengteng and Liu, Zhe and Chen, Xiwu and Bai, Xiang},
  booktitle={European conference on computer vision},
  pages={35--52},
  year={2020},
  organization={Springer}
}

@inproceedings{li2023logonet,
  title={Logonet: Towards accurate 3d object detection with local-to-global cross-modal fusion},
  author={Li, Xin and Ma, Tao and Hou, Yuenan and Shi, Botian and Yang, Yuchen and Liu, Youquan and Wu, Xingjiao and Chen, Qin and Li, Yikang and Qiao, Yu and others},
  booktitle={Proceedings of the IEEE/CVF conference on computer vision and pattern recognition},
  pages={17524--17534},
  year={2023}
}

@article{wu2023mvfusion,
  title={Mvfusion: Multi-view 3d object detection with semantic-aligned radar and camera fusion},
  author={Wu, Zizhang and Chen, Guilian and Gan, Yuanzhu and Wang, Lei and Pu, Jian},
  journal={arXiv preprint arXiv:2302.10511},
  year={2023}
}

@INPROCEEDINGS{Song2024LiRaFusion,
  author={Song, Jingyu and Zhao, Lingjun and Skinner, Katherine A.},
  booktitle={2024 IEEE International Conference on Robotics and Automation (ICRA)}, 
  title={LiRaFusion: Deep Adaptive LiDAR-Radar Fusion for 3D Object Detection}, 
  year={2024},
  volume={},
  number={},
  pages={18250-18257}}

@INPROCEEDINGS{Wolters2025UnleashingHydra,
  author={Wolters, Philipp and Gilg, Johannes and Teepe, Torben and Herzog, Fabian and Laouichi, Anouar and Hofmann, Martin and Rigoll, Gerhard},
  booktitle={2025 IEEE International Conference on Robotics and Automation (ICRA)}, 
  title={Unleashing HyDRa: Hybrid Fusion, Depth Consistency and Radar for Unified 3D Perception}, 
  year={2025},
  volume={},
  number={},
  pages={7467-7474}}

@article{Qi2026FusionBevLA,
  title={FusionBev: LiDAR and 4D radar fusion for 3D object detection},
  author={Yuan Xiao Qi and Chun Liu and Hangbin Wu and Ruijie Chen and Chenglu Wen and Xun Huang and Shoujun Jia and Keke Zhang},
  journal={Inf. Fusion},
  year={2026},
  volume={132},
  pages={104240}
}

@INPROCEEDINGS{Qian2021RobustFoggy,
  author={Qian, Kun and Zhu, Shilin and Zhang, Xinyu and Li, Li Erran},
  booktitle={2021 IEEE/CVF Conference on Computer Vision and Pattern Recognition (CVPR)}, 
  title={Robust Multimodal Vehicle Detection in Foggy Weather Using Complementary Lidar and Radar Signals}, 
  year={2021},
  volume={},
  number={},
  pages={444-453}}

@article{Hahner2021FogSO,
  title={Fog Simulation on Real LiDAR Point Clouds for 3D Object Detection in Adverse Weather},
  author={Martin Hahner and Christos Sakaridis and Dengxin Dai and Luc Van Gool},
  journal={2021 IEEE/CVF International Conference on Computer Vision (ICCV)},
  year={2021},
  pages={15263-15272}
}

@article{Kong2023Robo3DTR,
  title={Robo3D: Towards Robust and Reliable 3D Perception against Corruptions},
  author={Lingdong Kong and You-Chen Liu and Xin Li and Runnan Chen and Wenwei Zhang and Jiawei Ren and Liang Pan and Kaili Chen and Ziwei Liu},
  journal={2023 IEEE/CVF International Conference on Computer Vision (ICCV)},
  year={2023},
  pages={19937-19949}
}

@inproceedings{Chae2024LiDARBasedA3,
  title={LiDAR-Based All-Weather 3D Object Detection via Prompting and Distilling 4D Radar},
  author={Yujeong Chae and Hyeonseong Kim and Chang-Hwan Oh and Minseok Kim and Kuk-Jin Yoon},
  booktitle={European Conference on Computer Vision},
  year={2024}
}

@misc{wu2026unida3dunifieddomainadaptiveframework,
      title={UniDA3D: A Unified Domain-Adaptive Framework for Multi-View 3D Object Detection}, 
      author={Hongjing Wu and Cheng Chi and Jinlin Wu and Yanzhao Su and Zhen Lei and Wenqi Ren},
      year={2026},
      eprint={2603.27995},
      archivePrefix={arXiv},
      primaryClass={cs.CV}
}

@INPROCEEDINGS{Yang2021ST3D,
  author={Yang, Jihan and Shi, Shaoshuai and Wang, Zhe and Li, Hongsheng and Qi, Xiaojuan},
  booktitle={2021 IEEE/CVF Conference on Computer Vision and Pattern Recognition (CVPR)}, 
  title={ST3D: Self-training for Unsupervised Domain Adaptation on 3D Object Detection}, 
  year={2021},
  volume={},
  number={},
  pages={10363-10373}}

@article{kong2023rtnh+,
  title={RTNH+: Enhanced 4D radar object detection network using combined CFAR-based two-level preprocessing and vertical encoding},
  author={Kong, Seung-Hyun and Paek, Dong-Hee and Cho, Sangjae},
  journal={arXiv preprint arXiv:2310.17659},
  year={2023}
}

@inproceedings{wang2023bi,
  title={Bi-lrfusion: Bi-directional lidar-radar fusion for 3d dynamic object detection},
  author={Wang, Yingjie and Deng, Jiajun and Li, Yao and Hu, Jinshui and Liu, Cong and Zhang, Yu and Ji, Jianmin and Ouyang, Wanli and Zhang, Yanyong},
  booktitle={Proceedings of the IEEE/CVF Conference on Computer Vision and Pattern Recognition},
  pages={13394--13403},
  year={2023}
}

@article{xiong2023lxl,
  title={LXL: LiDAR excluded lean 3D object detection with 4D imaging radar and camera fusion},
  author={Xiong, Weiyi and Liu, Jianan and Huang, Tao and Han, Qing-Long and Xia, Yuxuan and Zhu, Bing},
  journal={IEEE Transactions on Intelligent Vehicles},
  volume={9},
  number={1},
  pages={79--92},
  year={2023},
  publisher={IEEE}
}

@inproceedings{radford2021clip,
  title={Learning transferable visual models from natural language supervision},
  author={Radford, Alec and Kim, Jong Wook and Hallacy, Chris and Ramesh, Aditya and Goh, Gabriel and Agarwal, Sandhini and Sastry, Girish and Askell, Amanda and Mishkin, Pamela and Clark, Jack and others},
  booktitle={International conference on machine learning},
  pages={8748--8763},
  year={2021},
  organization={PmLR}
}

@article{graham2015sparse,
  title={Sparse 3D convolutional neural networks},
  author={Graham, Ben},
  journal={arXiv preprint arXiv:1505.02890},
  year={2015}
}

@inproceedings{wu2023virtual,
  title={Virtual sparse convolution for multimodal 3d object detection},
  author={Wu, Hai and Wen, Chenglu and Shi, Shaoshuai and Li, Xin and Wang, Cheng},
  booktitle={Proceedings of the IEEE/CVF conference on computer vision and pattern recognition},
  pages={21653--21662},
  year={2023}
}

@inproceedings{xia2024hinted,
  title={Hinted: Hard instance enhanced detector with mixed-density feature fusion for sparsely-supervised 3d object detection},
  author={Xia, Qiming and Ye, Wei and Wu, Hai and Zhao, Shijia and Xing, Leyuan and Huang, Xun and Deng, Jinhao and Li, Xin and Wen, Chenglu and Wang, Cheng},
  booktitle={Proceedings of the IEEE/CVF Conference on Computer Vision and Pattern Recognition},
  pages={15321--15330},
  year={2024}
}

@inproceedings{xu2021spg,
  title={Spg: Unsupervised domain adaptation for 3d object detection via semantic point generation},
  author={Xu, Qiangeng and Zhou, Yin and Wang, Weiyue and Qi, Charles R and Anguelov, Dragomir},
  booktitle={Proceedings of the IEEE/CVF international conference on computer vision},
  pages={15446--15456},
  year={2021}
}

@article{yan2023cross,
  title={Cross modal transformer via coordinates encoding for 3d object dectection},
  author={Yan, Junjie and Liu, Yingfei and Sun, Jianjian and Jia, Fan and Li, Shuailin and Wang, Tiancai and Zhang, Xiangyu},
  journal={arXiv preprint arXiv:2301.01283},
  volume={2},
  number={3},
  pages={4},
  year={2023},
  publisher={ArXiv}
}

@article{do2022lossdistillnet,
  title={LossDistillNet: 3D object detection in point cloud under harsh weather conditions},
  author={Do, Anh The and Yoo, Myungsik},
  journal={IEEE Access},
  volume={10},
  pages={84882--84893},
  year={2022},
  publisher={IEEE}
}

@inproceedings{wang2023ssda3d,
  title={Ssda3d: Semi-supervised domain adaptation for 3d object detection from point cloud},
  author={Wang, Yan and Yin, Junbo and Li, Wei and Frossard, Pascal and Yang, Ruigang and Shen, Jianbing},
  booktitle={Proceedings of the AAAI Conference on Artificial Intelligence},
  volume={37},
  number={3},
  pages={2707--2715},
  year={2023}
}

@article{hao2025mimo,
  title={Mimo-embodied: X-embodied foundation model technical report},
  author={Hao, Xiaoshuai and Zhou, Lei and others},
  journal={arXiv preprint arXiv:2511.16518},
  year={2025}
}

@inproceedings{hao2023mixgen,
  title={Mixgen: A new multi-modal data augmentation},
  author={Hao, Xiaoshuai and Zhu, Yi and Appalaraju, Srikar and Zhang, Aston and Zhang, Wanqian and Li, Bo and Li, Mu},
  booktitle={Proceedings of the IEEE/CVF winter conference on applications of computer vision},
  pages={379--389},
  year={2023}
}

@inproceedings{hao2024your,
  title={Is Your HD Map Constructor Reliable under Sensor Corruptions?},
  author={Hao, Xiaoshuai and Wei, Mengchuan and Yang, Yifan and Zhao, Haimei and Zhang, Hui and Zhou, Yi and Wang, Qiang and Li, Weiming and Kong, Lingdong and Zhang, Jing},
  booktitle={Advances in Neural Information Processing System},
  year={2024}
}

@inproceedings{hao2024mapdistill,
  title={MapDistill: Boosting Efficient Camera-based HD Map Construction via Camera-LiDAR Fusion Model Distillation},
  author={Hao, Xiaoshuai and Li, Ruikai and Zhang, Hui and Li, Dingzhe and Yin, Rong and Jung, Sangil and Park, Seung-In and Yoo, ByungIn and Zhao, Haimei and Zhang, Jing},
  booktitle={European Conference on Computer Vision},
  year={2024}
}

@article{hao2025mapfusion,
  title={Mapfusion: A novel bev feature fusion network for multi-modal map construction},
  author={Hao, Xiaoshuai and Diao, Yunfeng and Wei, Mengchuan and Yang, Yifan and Hao, Peng and Yin, Rong and Zhang, Hui and Li, Weiming and Zhao, Shu and Liu, Yu},
  journal={Information Fusion},
  volume={119},
  pages={103018},
  year={2025},
  publisher={Elsevier}
}

@inproceedings{hao2025msc,
  title={Msc-bench: Benchmarking and analyzing multi-sensor corruption for driving perception},
  author={Hao, Xiaoshuai and Liu, Guanqun and Zhao, Yuting and Ji, Yuheng and Wei, Mengchuan and Zhao, Haimei and Kong, Lingdong and Yin, Rong and Liu, Yu},
  booktitle={2025 IEEE International Conference on Multimedia and Expo (ICME)},
  pages={1--6},
  year={2025},
  organization={IEEE}
}

@inproceedings{hao2025really,
  title={What Really Matters for Robust Multi-Sensor HD Map Construction?},
  author={Hao, Xiaoshuai and Zhao, Yuting and Ji, Yuheng and Dai, Luanyuan and Hao, Peng and Li, Dingzhe and Cheng, Shuai and Yin, Rong},
  booktitle={2025 IEEE/RSJ International Conference on Intelligent Robots and Systems (IROS)},
  pages={1298--1304},
  year={2025},
  organization={IEEE}
}

@article{shan2025stability,
  title={Stability under scrutiny: Benchmarking representation paradigms for online hd mapping},
  author={Shan, Hao and Li, Ruikai and Jiang, Han and others},
  journal={arXiv preprint arXiv:2510.10660},
  year={2025}
}

@inproceedings{hao2025safemap,
  title={SafeMap: Robust HD Map Construction from Incomplete Observations},
  author={Hao, Xiaoshuai and Kong, Lingdong and Yin, Rong and Wang, Pengwei and Zhang, Jing and Diao, Yunfeng and Zhao, Shu},
  booktitle={International Conference on Machine Learning},
  pages={22091--22102},
  year={2025},
  organization={PMLR}
}

@inproceedings{zhang2025mapnav,
  title={Mapnav: A novel memory representation via annotated semantic maps for vlm-based vision-and-language navigation},
  author={Zhang, Lingfeng and Hao, Xiaoshuai and Xu, Qinwen and Zhang, Qiang and Zhang, Xinyao and Wang, Pengwei and Zhang, Jing and Wang, Zhongyuan and Zhang, Shanghang and Xu, Renjing},
  booktitle={Proceedings of the 63rd Annual Meeting of the Association for Computational Linguistics (Volume 1: Long Papers)},
  pages={13032--13056},
  year={2025}
}

@article{zhang2025nava,
  title={$NavA^{3}$: Understanding Any Instruction, Navigating Anywhere, Finding Anything},
  author={Zhang, Lingfeng and Hao, Xiaoshuai and Tang, Yingbo and Fu, Haoxiang and Zheng, Xinyu and Wang, Pengwei and Wang, Zhongyuan and Ding, Wenbo and Zhang, Shanghang},
  journal={arXiv preprint arXiv:2508.04598},
  year={2025}
}

@inproceedings{tang2025roboafford,
  title={Roboafford: A dataset and benchmark for enhancing object and spatial affordance learning in robot manipulation},
  author={Tang, Yingbo and Zhang, Lingfeng and Zhang, Shuyi and Zhao, Yinuo and Hao, Xiaoshuai},
  booktitle={Proceedings of the 33rd ACM International Conference on Multimedia},
  pages={12706--12713},
  year={2025}
}

@article{hao2025roboafford++,
  title={RoboAfford++: A Generative AI-Enhanced Dataset for Multimodal Affordance Learning in Robotic Manipulation and Navigation},
  author={Hao, Xiaoshuai and Tang, Yingbo and Zhang, Lingfeng and Ma, Yanbiao and Diao, Yunfeng and Jia, Ziyu and Ding, Wenbo and Ye, Hangjun and Chen, Long},
  journal={arXiv preprint arXiv:2511.12436},
  year={2025}
}

@article{zhang2025your,
  title={Is your VLM Sky-Ready? A Comprehensive Spatial Intelligence Benchmark for UAV Navigation},
  author={Zhang, Lingfeng and Zhang, Yuchen and Li, Hongsheng and Fu, Haoxiang and Tang, Yingbo and Ye, Hangjun and Chen, Long and Liang, Xiaojun and Hao, Xiaoshuai and Ding, Wenbo},
  journal={arXiv preprint arXiv:2511.13269},
  year={2025}
}

@article{zhang2025socialnav,
  title={SocialNav-Map: Dynamic Mapping with Human Trajectory Prediction for Zero-Shot Social Navigation},
  author={Zhang, Lingfeng and Xiao, Erjia and Hao, Xiaoshuai and Fu, Haoxiang and Gong, Zeying and Chen, Long and Liang, Xiaojun and Xu, Renjing and Ye, Hangjun and Ding, Wenbo},
  journal={arXiv preprint arXiv:2511.12232},
  year={2025}
}

@article{zhang2026you,
  title={What You See is What You Reach: Towards Spatial Navigation with High-Level Human Instructions},
  author={Zhang, Lingfeng and Fu, Haoxiang and Hao, Xiaoshuai and Zhang, Shuyi and Zhang, Qiang and Liu, Rui and Chen, Long and Ding, Wenbo},
  year={2026}
}

@inproceedings{zhang2025video,
  title={Video-cot: A comprehensive dataset for spatiotemporal understanding of videos based on chain-of-thought},
  author={Zhang, Shuyi and Hao, Xiaoshuai and Tang, Yingbo and Zhang, Lingfeng and Wang, Pengwei and Wang, Zhongyuan and Ma, Hongxuan and Zhang, Shanghang},
  booktitle={Proceedings of the 33rd ACM International Conference on Multimedia},
  pages={12745--12752},
  year={2025}
}

@article{zhang2025team,
  title={Team Xiaomi EV-AD VLA: Caption-Guided Retrieval System for Cross-Modal Drone Navigation--Technical Report for IROS 2025 RoboSense Challenge Track 4},
  author={Zhang, Lingfeng and Xiao, Erjia and Zhang, Yuchen and Fu, Haoxiang and Hu, Ruibin and Ma, Yanbiao and Ding, Wenbo and Chen, Long and Ye, Hangjun and Hao, Xiaoshuai},
  journal={arXiv preprint arXiv:2510.02728},
  year={2025}
}

@article{xiao2025team,
  title={Team Xiaomi EV-AD VLA: Learning to Navigate Socially Through Proactive Risk Perception--Technical Report for IROS 2025 RoboSense Challenge Social Navigation Track},
  author={Xiao, Erjia and Zhang, Lingfeng and Tang, Yingbo and Cheng, Hao and Xu, Renjing and Ding, Wenbo and Zhou, Lei and Chen, Long and Ye, Hangjun and Hao, Xiaoshuai},
  journal={arXiv e-prints},
  pages={arXiv--2510},
  year={2025}
}

@article{kong2026robosense,
  title={The RoboSense challenge: Sense anything, navigate anywhere, adapt across platforms},
  author={Kong, Lingdong and Xie, Shaoyuan and Gong, Zeying and Li, Ye and Chu, Meng and Liang, Ao and Dong, Yuhao and Hu, Tianshuai and Qiu, Ronghe and Li, Rong and others},
  journal={arXiv preprint arXiv:2601.05014},
  year={2026}
}

@article{zheng2025railway,
  title={Railway side slope hazard detection system based on generative models},
  author={Zheng, Xinyu and He, Yangfan and Luo, Yuhao and Zhang, Lingfeng and Wang, Jianhui and Shi, Tianyu and Bai, Yun},
  journal={IEEE Sensors Journal},
  year={2025},
  publisher={IEEE}
}

@inproceedings{zhang2024trihelper,
  title={Trihelper: Zero-shot object navigation with dynamic assistance},
  author={Zhang, Lingfeng and Zhang, Qiang and Wang, Hao and Xiao, Erjia and Jiang, Zixuan and Chen, Honglei and Xu, Renjing},
  booktitle={2024 IEEE/RSJ International Conference on Intelligent Robots and Systems (IROS)},
  pages={10035--10042},
  year={2024},
  organization={IEEE}
}

@inproceedings{zhang2025multi,
  title={Multi-floor zero-shot object navigation policy},
  author={Zhang, Lingfeng and Wang, Hao and Xiao, Erjia and Zhang, Xinyao and Zhang, Qiang and Jiang, Zixuan and Xu, Renjing},
  booktitle={2025 IEEE International Conference on Robotics and Automation (ICRA)},
  pages={6416--6422},
  year={2025},
  organization={IEEE}
}

@article{gong2025stairway,
  title={Stairway to success: Zero-shot floor-aware object-goal navigation via llm-driven coarse-to-fine exploration},
  author={Gong, Zeying and Li, Rong and Hu, Tianshuai and Qiu, Ronghe and Kong, Lingdong and Zhang, Lingfeng and Ding, Yiyi and Zhang, Leying and Liang, Junwei},
  journal={arXiv e-prints},
  pages={arXiv--2505},
  year={2025}
}

@article{zhang2025humanoidpano,
  title={Humanoidpano: Hybrid spherical panoramic-lidar cross-modal perception for humanoid robots},
  author={Zhang, Qiang and Zhang, Zhang and Cui, Wei and Sun, Jingkai and Cao, Jiahang and Guo, Yijie and Han, Gang and Zhao, Wen and Wang, Jiaxu and Sun, Chenghao and others},
  journal={arXiv preprint arXiv:2503.09010},
  year={2025}
}

@article{wu2025evaluating,
  title={Evaluating GPT-4o's Embodied Intelligence: A Comprehensive Empirical Study},
  author={Wu, Yujie and Lyu, Huaihai and Tang, Yingbo and Zhang, Lingfeng and Zhang, Zhihui and Zhou, Wei and Hao, Siqi},
  journal={Authorea Preprints},
  year={2025},
  publisher={Authorea}
}

@article{liu2025toponav,
  title={Toponav: Topological graphs as a key enabler for advanced object navigation},
  author={Liu, Peiran and Zhang, Qiang and Peng, Daojie and Zhang, Lingfeng and Qin, Yihao and Zhou, Hang and Ma, Jun and Xu, Renjing and Ji, Yiding},
  journal={arXiv preprint arXiv:2509.01364},
  year={2025}
}

@inproceedings{li2025vquala,
  title={Vquala 2025 challenge on engagement prediction for short videos: Methods and results},
  author={Li, Dasong and Ma, Sizhuo and Hua, Hang and Li, Wenjie and Wang, Jian and Zhou, Chris Wei and Guan, Fengbin and Li, Xin and Yu, Zihao and Lu, Yiting and others},
  booktitle={Proceedings of the IEEE/CVF International Conference on Computer Vision},
  pages={3391--3401},
  year={2025}
}

@article{cheng2025exploring,
  title={Exploring typographic visual prompts injection threats in cross-modality generation models},
  author={Cheng, Hao and Xiao, Erjia and Wang, Yichi and Zhang, Lingfeng and Zhang, Qiang and Cao, Jiahang and Xu, Kaidi and Sun, Mengshu and Hao, Xiaoshuai and Gu, Jindong and others},
  journal={arXiv preprint arXiv:2503.11519},
  year={2025}
}

@article{zhang2025lips,
  title={Lips: Large-scale humanoid robot reinforcement learning with parallel-series structures},
  author={Zhang, Qiang and Han, Gang and Sun, Jingkai and Zhao, Wen and Cao, Jiahang and Wang, Jiaxu and Cheng, Hao and Zhang, Lingfeng and Guo, Yijie and Xu, Renjing},
  journal={arXiv preprint arXiv:2503.08349},
  year={2025}
}

@article{hao2026h2r,
  title={H2R-BM: Can Leveraging Human Videos Enhance Performance and Generalizability in Robotic Bimanual Manipulation?},
  author={Hao, Xiaoshuai and Lyu, Huaihai and Zhang, Lingfeng and Liu, Rui and Wu, Dayan and Zhang, Jing and Chen, Long},
  journal={Pattern Recognition},
  pages={113637},
  year={2026},
  publisher={Elsevier}
}

@article{fu2026sef,
  title={SEF-MAP: Subspace-Decomposed Expert Fusion for Robust Multimodal HD Map Prediction},
  author={Fu, Haoxiang and Zhang, Lingfeng and Li, Hao and Hu, Ruibing and Li, Zhengrong and Liu, Guanjing and Tan, Zimu and Chen, Long and Ye, Hangjun and Hao, Xiaoshuai},
  journal={arXiv preprint arXiv:2602.21589},
  year={2026}
}

@article{zhang2026meshmimic,
  title={MeshMimic: Geometry-Aware Humanoid Motion Learning through 3D Scene Reconstruction},
  author={Zhang, Qiang and Ma, Jiahao @inproceedings{hao2024mbfusion,
  title={Mbfusion: A new multi-modal bev feature fusion method for hd map construction},
  author={Hao, Xiaoshuai and Zhang, Hui and Yang, Yifan and Zhou, Yi and Jung, Sangil and Park, Seung-In and Yoo, ByungIn},
  booktitle={2024 IEEE International Conference on Robotics and Automation (ICRA)},
  pages={15922--15928},
  year={2024},
  organization={IEEE}
}and Liu, Peiran and Shi, Shuai and Su, Zeran and Wang, Zifan and Sun, Jingkai and Cui, Wei and Yu, Jialin and Han, Gang and others},
  journal={arXiv preprint arXiv:2602.15733},
  year={2026}
}

@inproceedings{hao2024mbfusion,
  title={Mbfusion: A new multi-modal bev feature fusion method for hd map construction},
  author={Hao, Xiaoshuai and Zhang, Hui and Yang, Yifan and Zhou, Yi and Jung, Sangil and Park, Seung-In and Yoo, ByungIn},
  booktitle={2024 IEEE International Conference on Robotics and Automation (ICRA)},
  pages={15922--15928},
  year={2024},
  organization={IEEE}
}


\end{document}